\documentclass{vgtc}

\onlineid{1320}

\vgtccategory{Research}

\title{Vector Field Synthesis with Sparse Streamlines Using Diffusion Model}

\author{Nguyen K.\ Phan\thanks{e-mail:nguyenpkk95@gmail.com} %
\and Ricardo \ Morales \thanks{e-mail:ramoral4@cougarnet.uh.edu}
\and Sebastian D. \ Espriella \thanks{e-mail:sdelaesp@cougarnet.uh.edu}
\and Guoning Chen\thanks{e-mail:gchen22@central.uh.edu}}
\affiliation{\scriptsize University of Houston}

\abstract{%
We present a novel diffusion-based framework for synthesizing 2D vector fields from sparse, coherent inputs (i.e., streamlines) while maintaining physical plausibility. Our method employs a conditional denoising diffusion probabilistic model with classifier-free guidance, enabling progressive reconstruction that preserves both geometric and physical constraints. 
Experimental results demonstrate our method's ability to synthesize plausible vector fields that adhere to physical laws while maintaining fidelity to sparse input observations, outperforming traditional optimization-based approaches in terms of flexibility and physical consistency. 

}


\graphicspath{{figs/}{figures/}{pictures/}{images/}{./}} 

\usepackage{mathptmx}                  
\usepackage{bm} 
\usepackage{float}

\usepackage{algorithm}
\usepackage{algpseudocode}
\usepackage{multirow}

\usepackage{amsmath,amssymb}

\begin{document}


\setlength{\baselineskip}{0.99 \baselineskip}
\setlength{\abovedisplayskip}{1pt}
\setlength{\belowdisplayskip}{1pt}
\setlength{\abovedisplayshortskip}{1pt}
\setlength{\belowdisplayshortskip}{-2pt}
\setlength{\belowcaptionskip}{2pt}
\setlength{\abovecaptionskip}{1pt}
\setlength{\textfloatsep}{1pt}
\setlength{\floatsep}{1pt}
\setlength{\intextsep}{1pt}

\maketitle

\section{Introduction}

Vector fields are essential for understanding complex dynamical systems (e.g., turbulence) in scientific research and engineering applications. However, these fields often contain intricate features that are difficult to analyze. While progress has been made in characterizing these features \cite{adeelhairpin2023,zafar2024topological}, experts need a framework that allows direct control over the presence and properties of relevant features to reduce the complexity caused by unwanted features. Vector field synthesis addresses this need by generating smooth vector fields from user-specified constraints.

Existing vector field synthesis techniques are mainly formulated as specific constrained optimization problems~\cite{DirFieldSIGA2016, VFProcessingSIG2016Course}. While they usually produce continuous and smooth vector fields that are sufficient for various graphics, robotics, and visualization applications \cite{DirFieldSIGA2016,lindemann2005smoothly,kapitanyuk2017guiding,chen07,Zhang06,TimeVaryingDesign12,wang2016flowvisual}, they often do not guarantee the physical authenticity of the produced fields. 
More importantly, these techniques often generate overly simplified patterns in unconstrained regions due to energy minimization, limiting their utility for exploring diverse flow patterns needed by experts' study of complex features and their relations.

In the meantime, reconstruction and upsampling (or super resolution) techniques have been proposed to address the gap between the needs of ultra-detailed data for knowledge discovery and the lack of disk space to store high-resolution data. While sharing some similarities to vector field synthesis, reconstruction is different from synthesis. First, reconstruction usually has ground truth, while synthesis often doesn't. Second, the samples used to produce low-res data, later for upsampling or reconstructions, are usually carefully placed and are consistent (in terms of the number and distribution of samples) for the same type of flows. In contrast, the input constraints for vector field synthesis are vastly different for different goals, and are often sparse. That said, while not necessarily more challenging than reconstruction, vector field synthesis requires different techniques to address its unique challenges.

On the other hand, vector field synthesis shares similarities with image inpainting, where missing regions are filled based on surrounding information. Denoising diffusion probabilistic models (DDPMs) \cite{ho2020denoising} have shown success in inpainting tasks by learning underlying data distributions. This inspires our approach. In this work, we propose to train a \emph{conditional diffusion model} for vector field synthesis given sparse streamlines. 
We utilize a sparse conditioning mechanism that maintains fidelity to observed streamlines while allowing for physical consistency in unobserved regions. 
We propose a classifier-free guidance approach that balances adherence to streamline constraints with physical plausibility. Unlike traditional approaches that rely on energy minimization, our diffusion-based method learns the underlying distribution of physically plausible vector fields from training data. This enables our model to generate diverse yet physically consistent vector fields even in regions far from observed streamlines. We apply our method to synthesize 2D vector fields with various complexities.

\section{Related Work}
\label{sec:related_work}

\textbf{Classic vector field design and synthesis} allow the generation of continuous vector fields from user-specified constraints. 
It has been applied to various computer graphics applications~\cite{Turk:01,Wei:01,stam:99:stable,Stam:2003:FSAT,Funck:06:vfDeform,Fu:07:hair} and development of visualization techniques~\cite{vanWijk:2003:IBFVS,chen07,wang2016flowvisual}. 
Among existing methods, topology-based techniques have been widely applied \cite{Zhang06,Fisher:07:Tangent,chen07}, with extensions to tensor field synthesis \cite{zhang2006interactive,palacios2016tensor} and N-way rotational symmetric field design \cite{Palacios:ROSY:07,Ray:DGF:08,Lai10}. 
Chen et al.~\cite{TimeVaryingDesign12} extended the topology-based approach to the design of unsteady flows. 
These methods focus primarily on the synthesis of geometric characteristics of the flow rather than its physical properties.






\vspace{0.02in}
\noindent\textbf{Vector field reconstruction and upsampling} aim to recover or enhance vector fields from sparse or low-resolution data. 
Traditional reconstruction approaches include interpolation methods \cite{lindemann2005smoothly} and physics-based reconstruction \cite{kapitanyuk2017guiding}. 

Recently, machine-learning based methods have received increasing attention. 
Page \cite{page2025super} achieved super-resolution of turbulence using deep neural networks to predict high-resolution flow from coarse vorticity observations. 
Shu et al.~\cite{shu2023physics} presented a physics-informed diffusion model for vorticity field reconstruction. They later extended it to addressing the reconstruction of the vorticity field within the uniformly distributed square regions. 
Our work differs from theirs as we train a diffusion model using velocity (vector) values directly rather than separating them into multiple scalar quantities. Furthermore, our constraints are given along individual streamlines rather than in fixed square regions.

Gu et al.~\cite{gu2021reconstructing} proposed VFR-UFD, a deep learning framework for reconstructing unsteady flow data from representative streamlines. VFR-UFD differs from our approach in that it aims to recover known vector fields rather than synthesize new, physically plausible fields from minimal constraints.  Our work addresses the challenging task of true synthesis rather than reconstruction, moving beyond recovering existing fields to generating entirely new ones that satisfy both geometric constraints and physical plausibility requirements.

\vspace{0.02in}
\noindent\textbf{Physics-Informed Neural Networks (PINNs)}~\cite{raissi2019physics} incorporate physical laws into neural networks by encoding partial differential equations directly into loss functions. While effective for solving and discovering PDEs in specific physical systems, PINNs differ fundamentally from our approach. They are typically one-shot methods designed for individual problem instances rather than general synthesis tasks, and cannot readily generate diverse vector fields from sparse inputs like streamlines.

\section{Diffusion Model for Vector Field Synthesis}

\begin{figure}[htbp]
    \centering
    \includegraphics[width=\columnwidth]{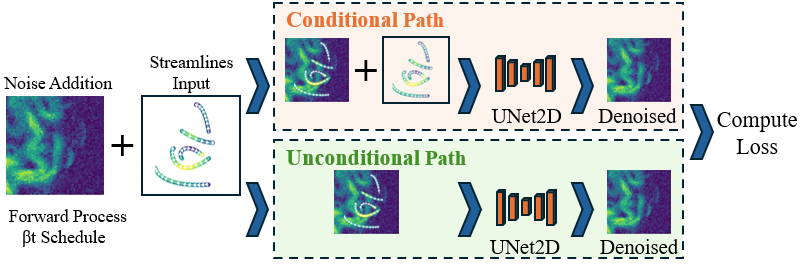}
    \caption{Overview of our vector field diffusion training pipeline. The process begins with noise addition following a forward diffusion process with $\beta_t$ schedule, combined with streamline input values. Both conditional and unconditional paths through parallel UNet2D networks are modeled, with the outputs combined using classifier-free guidance (CFG). This dual-path architecture enables the model to balance fidelity to known streamline regions while generating physically plausible completions in unknown regions.}
    \label{fig:training_pipeline}
\end{figure}

\subsection{Model Architecture}
We employ a conditional UNet architecture adapted for vector field data with 4 input channels (2 for noisy vector field components and 2 for mask channels) and 2 output channels (predicted noise for vector field components). The network consists of 4 downsampling and upsampling blocks.

\subsection{Vector Field Diffusion Process}
Our approach builds upon the denoising diffusion probabilistic model framework, adapting it specifically for vector field data. The forward diffusion process gradually adds noise according to a predefined $\beta_t$ schedule as shown in Figure~\ref{fig:training_pipeline}. This process transforms smooth vector fields into progressively noisier versions while preserving the known values along streamlines. 
Our implementation ensures that the noise addition respects the physical constraints of the vector field.

\subsection{Classifier-Free Guidance for Vector Fields}
To enhance quality and controllability, we incorporate a Classifier-Free Guidance (CFG) within our diffusion framework. As illustrated in Figure~\ref{fig:training_pipeline}, our approach explicitly models both conditional and unconditional paths through parallel UNet2D networks:

\begin{enumerate}[leftmargin=*]
  \setlength{\itemsep}{0pt}
  \setlength{\parskip}{0pt}
    \item During training, we randomly drop the conditioning information (mask $\mathbf{M}$) with probability $p_{\text{drop}} = 0.1$ to create both conditional and unconditional samples
    \item The model processes inputs through two parallel paths:
      \begin{itemize}[leftmargin=3pt]
        \item Conditional path: UNet2D receives the noisy vector field and the streamline mask
        \item Unconditional path: UNet2D receives the noisy vector field without mask conditioning
      \end{itemize}
    \item The outputs from both paths are combined as:
    \begin{equation}
    \hat{\bm{\epsilon}}_{\text{CFG}} = \hat{\bm{\epsilon}}_{\text{uncond}} + w \cdot (\hat{\bm{\epsilon}}_{\text{cond}} - \hat{\bm{\epsilon}}_{\text{uncond}})
    \end{equation}
\end{enumerate}

The guidance scale $w$ controls the influence of conditioning information. Higher values (typically 1-10) result in vector fields that more closely adhere to streamline constraints, while values closer to 1 allow for more diversity. We empirically determined that $w = 3.0$ provides an optimal balance between constraint fidelity and physical plausibility. This dual-path architecture, as shown in Figure~\ref{fig:training_pipeline}, is key to our method's ability to generate physically consistent vector fields that respect the provided streamline constraints.


\subsection{Streamline-Constrained Training}

We employ a uniform grid-based seeding strategy to generate $N$ streamlines using $4^{th}-$order Runge-Kutta integration with adaptive step sizes. These sparse streamlines are converted to a binary mask $\mathbf{M} \in \{0, 1\}^{H \times W}$ where $\mathbf{M}_{i,j} = 1$ indicates a point on or near a streamline.

Next, we combine noise addition with the obtained streamline marks (Figure~\ref{fig:training_pipeline}). This is fundamentally different from previous methods.
Recent works such as \cite{shu2023physics} and \cite{shu2024inpainting} train diffusion models to denoise completely noisy samples without preserving any specific values during the denoising process. In contrast, our method preserves ground-truth vector values along streamlines during training:
\begin{equation}
\mathbf{x}_{\text{noisy}} = \begin{cases}
\mathbf{x}_{\text{clean}} & \text{where } \mathbf{M} > 0.5 \\
\mathbf{x}_{\text{noisy}} & \text{otherwise}
\end{cases}
\end{equation}

This selective noise application teaches the model to respect known vector values while generating physically plausible vector fields in unknown regions. Critically, we compute the loss only in the unknown regions:
\begin{equation}
\mathcal{L} = \|\hat{\bm{\epsilon}} \cdot (1 - \mathbf{M}) - \bm{\epsilon} \cdot (1 - \mathbf{M})\|^2
\end{equation}
\noindent where $\hat{\bm{\epsilon}}$ is the predicted noise, $\bm{\epsilon}$ is the actual noise, and $(1 - \mathbf{M})$ masks out the known streamline regions. This masked loss computation is essential for training the model to focus on generating plausible vector field values in unknown regions while preserving the known values along streamlines.

This training strategy is crucial to ensure the synthesized field adheres to input streamlines while maintaining physical consistency in unknown regions.

\section{Sampling Process}

\begin{figure}[htbp]
    \centering
    \includegraphics[width=\columnwidth]{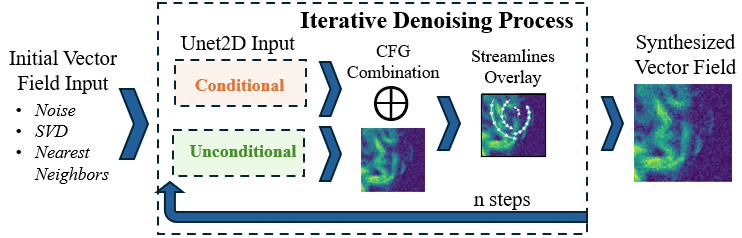}
    \caption{Our vector field diffusion inference pipeline. The process begins with random noise in unknown regions while preserving known vector data from streamline observations. Through iterative denoising steps guided by CFG, the model produces complete vector fields while maintaining physical consistency. 
    }
    \label{fig:inference_pipeline}
\end{figure}

During inference, we start with a known streamline mask $\mathbf{M}$ and corresponding vector values $\mathbf{x}_{\text{masked}} = \mathbf{M} \odot \mathbf{x}_0$. The unknown regions are initialized with random noise. The sampling process then proceeds as follows:
\begin{enumerate} [leftmargin=*]
  \setlength{\itemsep}{0pt}
  \setlength{\parskip}{0pt}
    \item Initialize: $\mathbf{x}_T = \mathbf{M} \odot \mathbf{x}_{\text{masked}} + (1-\mathbf{M}) \odot \mathbf{z}$ where $\mathbf{z} \sim \mathcal{N}(\mathbf{0}, \mathbf{I})$
    \item For $t = T, T-1, \ldots, 1$:
    \begin{itemize} [leftmargin=3pt]
        \item Predict noise: $\hat{\bm{\epsilon}} = \bm{\epsilon}_\theta(\mathbf{x}_t, t, \mathbf{M})$
        \item Compute denoised estimate using scheduler
        \item Update only unknown regions:
        \begin{equation}
        \mathbf{x}_{t-1} = \mathbf{M} \odot \mathbf{x}_{\text{masked}} + (1-\mathbf{M}) \odot \text{Scheduler}(\mathbf{x}_t, t, \hat{\bm{\epsilon}})
        \end{equation}
    \end{itemize}
    \item Return final reconstruction $\mathbf{x}_0$
\end{enumerate}

\begin{figure*}[t]
    \centering
    \includegraphics[width=0.98\textwidth]{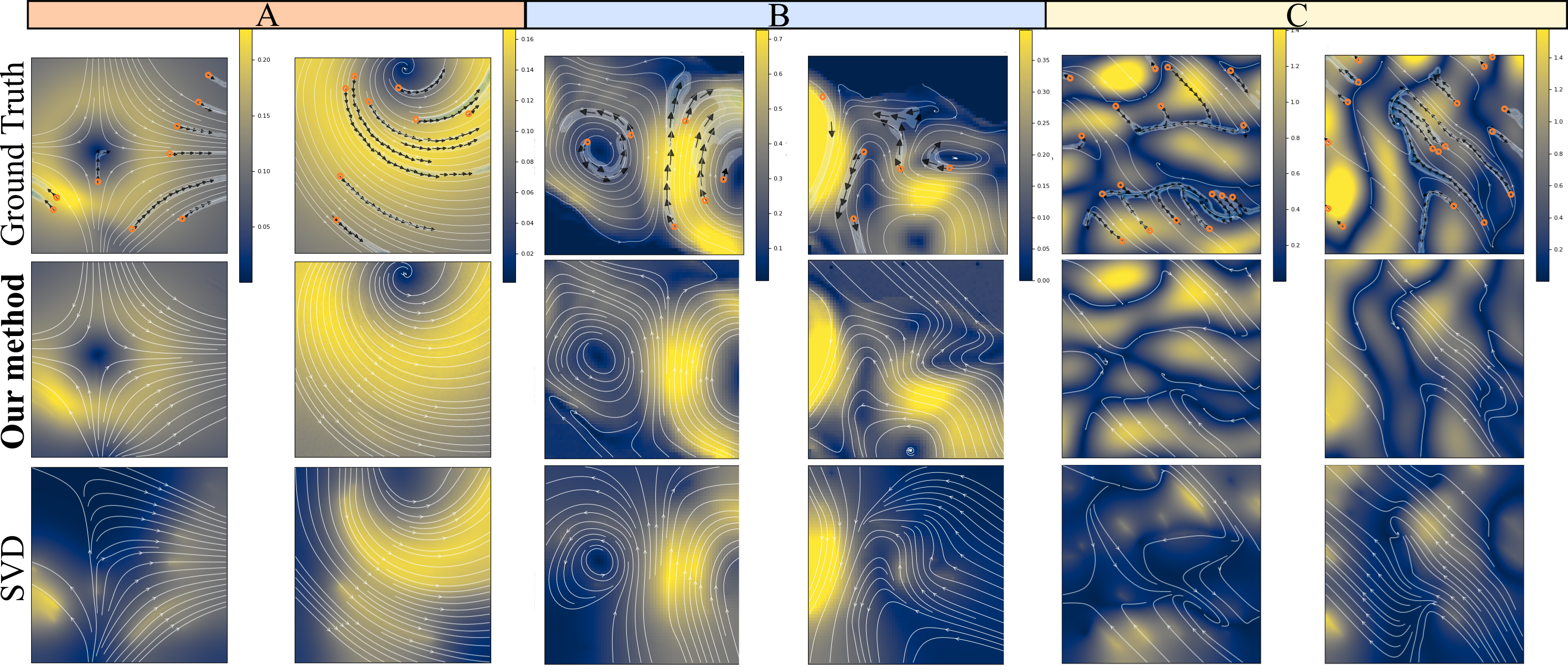}
    \caption{Qualitative comparison across all three datasets: \textbf{(A)} Synthetic, \textbf{(B)} Red Sea, and \textbf{(C)} CFD, with two examples per dataset. \textbf{Top row:} Ground truth vector fields with sparse streamline inputs overlaid that the model hasn't seen. \textbf{Middle row:} Vector fields synthesized by our diffusion model. \textbf{Bottom row:} Vector fields synthesized by the baseline optimization method. Our approach consistently outperforms the baseline, better capturing complex flow patterns and maintaining physical consistency throughout the domain.}
    \label{fig:combined_results}
\end{figure*}

This approach ensures that known streamline values remain fixed throughout sampling while unknown regions are progressively denoised based on the model's predictions. This key distinction from standard diffusion models maintains fidelity of observed data while generating physically plausible completions in unobserved regions. 
Without training the model to respect known values during the denoising process, it would not be able to generate coherent vector fields that seamlessly integrate with the provided streamline constraints.

\section{Experiments}
\label{sec:experiments}

To evaluate our diffusion-based approach, we conducted experiments using multiple datasets with varying characteristics. We compare our method against traditional optimization-based approaches.





\begin{figure}[htb]
    \centering
    \includegraphics[width=0.98\columnwidth]{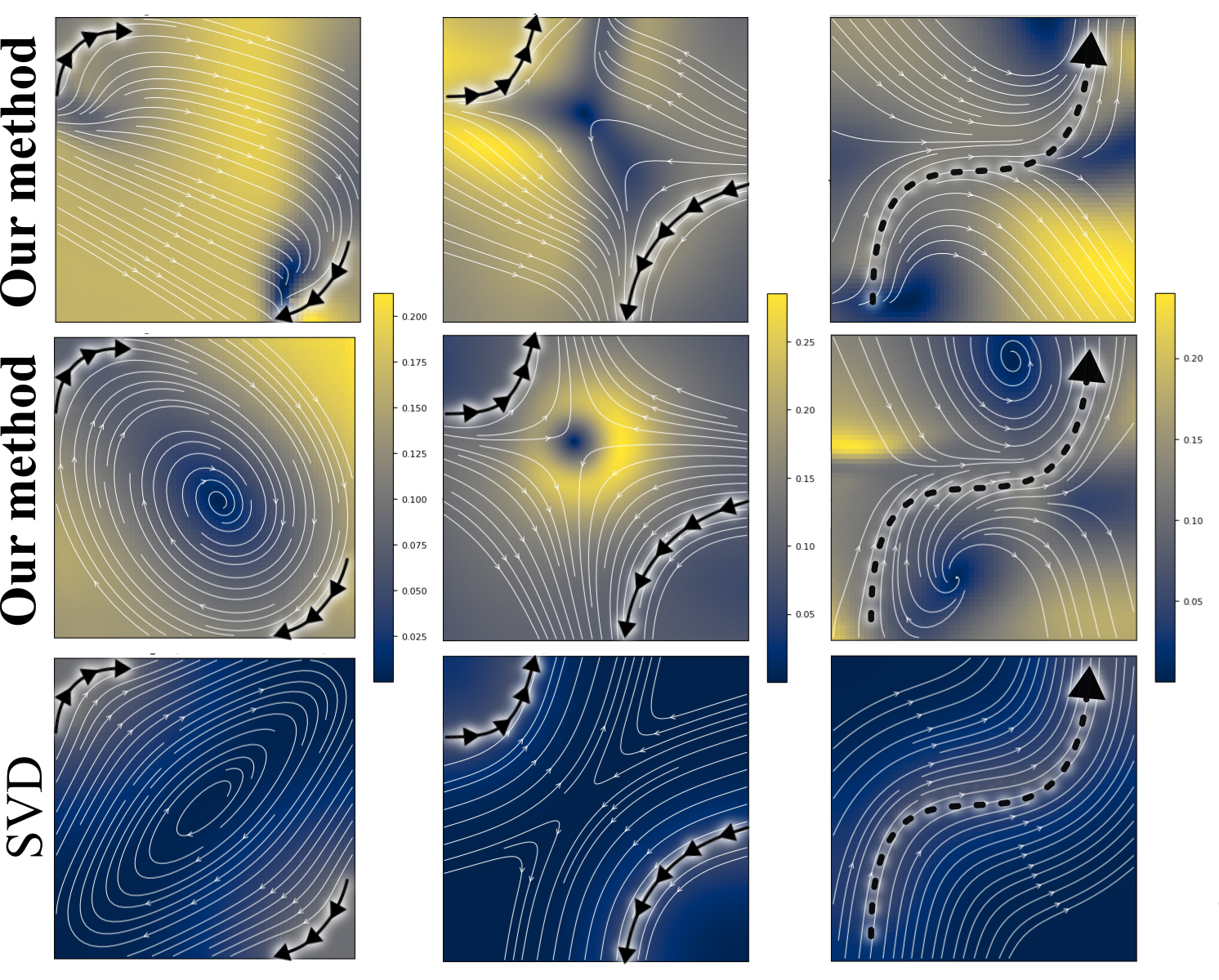}
    \caption{Vector field synthesis from hand-drawn streamlines. Each column shows a different input pattern: a simple vortex (left), a simple saddle (middle), and a complex S-shaped streamline (right).
    \textbf{Top row:} Results from our diffusion model with low CFG values and different random seeds, exhibiting more varied fields despite simple input streamlines. Note how the first column shows a vortex with near-parallel streamlines and the second column displays a more complex saddle pattern. \textbf{Middle row:} Results from our diffusion model with CFG $\geq$ 1, showing standard expected features that perfectly adhere to the input streamlines: a large simple vortex (left) and large simple saddle (middle). \textbf{Bottom row:} Results from the SVD method, which produces single, simple flows with physical properties concentrated around constraints for each set of constraints. The thick black lines with arrows are the hand-drawn input streamlines. The dashed black arrow signifies sparser sampling of a long hand-drawn streamline to smooth out artifacts like jitter or sharp angles from the drawing process. The thin white lines are streamlines depicting the synthesized flow.}
    \label{fig:synthesis_results}
\end{figure}

\subsection{Baseline Method: Streamline Vector Diffusion}
\label{subsec:baseline}

For our baseline, we implement a new Streamline Vector Diffusion (SVD), which adapts the concept of Gradient Vector Flow (GVF)~\cite{Turk:01} for vector field synthesis from streamlines. It achieves similar results as the constrained optimization \cite{chen07,Zhang06}. While GVF was designed for edge detection using gradient vectors, SVD directly diffuses vector information from sparse streamlines by minimizing:
\begin{equation}
E = \iint \mu(u_x^2 + u_y^2 + v_x^2 + v_y^2) + \psi(x,y)|\mathbf{v} - \mathbf{v}_0|^2 \, dx \, dy
\end{equation}

Key differences from GVF include: SVD (1) works with vector values directly from streamlines rather than gradient vectors, (2) uses a spatially varying weight function $\psi(x,y)$ optimized for streamline inputs, and (3) employs an iterative solver with explicit time-stepping. 
SVD is suitable for our comparison because, unlike upsampling methods~\cite{page2025super,shu2023physics} that expect uniformly distributed inputs, it handles the non-uniform nature of streamline constraints. Similar to Chen et al.~\cite{TimeVaryingDesign12}, SVD primarily focuses on directional information without incorporating a data-driven component that learns from physical simulations.

\subsection{Dataset Preparation}
\label{subsec:datasets}

We train our model separately on three different types of flows:
\begin{enumerate}[leftmargin=*]
  \setlength{\itemsep}{1pt}
  \setlength{\parskip}{1pt}
    \item \textbf{Synthetic flow:} We generated vector fields with known vortex boundaries using a parametric model based on Berenjkoub et al. \cite{berenjkoub2020vortex}. The vector field at any point $\mathbf{x} = (x, y)$ is generated using $\mathbf{v}(\mathbf{x}) = \mathbf{S}_i \cdot \mathbf{x} \cdot \frac{v_0(\|\mathbf{x}\|)}{\|\mathbf{x}\|}$, where $\mathbf{S}_i$ is one of three base shape matrices and $v_0(r)$ represents Vatistas' velocity profile. We created 41,000 distinct vector fields at $128 \times 128$ resolution.
    
    \item \textbf{Red Sea data:} This dataset consists of time-dependent 3D flow fields generated using the MIT ocean general circulation model (MITgcm), covering the Red Sea region (30°E-50°E, 10°N-30°N). We extracted approximately 3,800 2D vector field slices of size $128 \times 128$ from horizontal velocity components at multiple depths and time steps.
    
    \item \textbf{CFD data:} Our third dataset consists of two-dimensional Kolmogorov flow simulations from Shu et al. \cite{shu2023physics}, comprising numerical solutions to the incompressible Navier-Stokes equations with Reynolds number 1000. We processed 3,200 vector fields at $128 \times 128$ resolution.
\end{enumerate}
For our reconstruction experiments, we use a set of randomly seeded streamlines for each sample: 12 streamlines for the Synthetic and Red Sea datasets, and 20 for the CFD dataset. Some samples may have fewer streamlines, as a seed point may not produce a valid streamline, particularly in the low-velocity regions of the ocean data. All datasets were normalized to the range $[-1, 1]$ and split into training (90\%) and testing (10\%) sets.

\subsection{Results}
\label{subsec:results}

Although our primary goal is to synthesize new vector fields from user-defined streamlines, this task is inherently subjective and difficult to evaluate quantitatively. Therefore, to first validate our model's capabilities, we task it with \emph{reconstructing} known vector fields from a sparse set of streamlines. This reconstruction task 
provides an objective measure of how well our method can generate physically plausible fields that adhere to streamline constraints.

As shown in Figure~\ref{fig:combined_results}, our diffusion model effectively captures complex flow patterns across all three datasets. In the figure, the shaded areas along the streamlines indicate the masked regions where the ground-truth vector fields are sampled, and the color corresponds to vector magnitude. For the synthetic dataset (A), our model accurately reconstructs vortices and saddles. For the Red Sea dataset (B), it captures characteristic oceanographic patterns, including eddy structures and boundary currents. For the CFD dataset (C), it successfully reconstructs intricate turbulent structures characteristic of Kolmogorov flows. In contrast, the SVD baseline method produces overly smoothed fields in regions distant from constraints, failing to reconstruct intricate structures across all datasets. Additional results can be found in the supplemental document.

\begin{table}[h]
    \centering
    \footnotesize
    \begin{tabular}{llccc}
        \hline
        \textbf{Dataset} & \textbf{Method} & \textbf{MSE} $\downarrow$ & \textbf{Angular (°)} $\downarrow$ & \textbf{Physics} $\downarrow$ \\
        \hline
        \multirow{2}{*}{Synthetic} 
        & SVD & 0.005075 & 12.14 & 0.711280 \\
        & Ours & \textbf{0.000610} & \textbf{3.49} & \textbf{0.389157} \\
        \hline
        \multirow{2}{*}{Red Sea} 
        & SVD & 0.014640 & 37.16 & 0.653732 \\
        & Ours & \textbf{0.007494} & \textbf{25.61} & \textbf{0.425823} \\
        \hline
        \multirow{2}{*}{CFD} 
        & SVD & 0.144679 & 39.39 & 0.619762 \\
        & Ours & \textbf{0.038301} & \textbf{13.51} & \textbf{0.288306} \\
        \hline
    \end{tabular}
    \caption{Quantitative comparison across all datasets. Lower values indicate better performance.}
    \label{tab:combined_metrics}
\end{table}

Table~\ref{tab:combined_metrics} presents quantitative results comparing the methods across all three datasets. We evaluate using three metrics: Mean Squared Error (MSE) measuring overall vector field accuracy, Angular error measuring directional accuracy in degrees, and Physics error which combines normalized curl and divergence errors ($\frac{1}{2}(\frac{\|\nabla \times \mathbf{V} - \nabla \times \hat{\mathbf{V}}\|_2}{\|\nabla \times \mathbf{V}\|_2} + \frac{\|\nabla \cdot \mathbf{V} - \nabla \cdot \hat{\mathbf{V}}\|_2}{\|\nabla \cdot \mathbf{V}\|_2})$) to measure preservation of rotational and expansion/contraction characteristics.

Our approach consistently outperforms the SVD method across all metrics and datasets, with particularly notable improvements in physics error. This indicates better preservation of properties, such as curl and divergence, essential for physically plausible flow fields. The supplemental document compares the distributions of these physical properties between the reconstructed flows, using the two approaches, respectively, with the reference flows.

Our model was trained on an NVIDIA 3060 Ti GPU (16GB VRAM). Training took 22.7 hours for the Synthetic dataset, 3.47 hours for CFD, and 3.43 hours for Red Sea data. The longer time for Synthetic reflects its larger size (41,000 samples/epoch vs. ~3,200–3,800 for others). To stay within the GPU's 16GB VRAM limit, batch sizes were set to 48 for both the Synthetic and CFD datasets, and to 96 for the smaller Red Sea dataset. During inference, the model uses 600MB of VRAM, plus approximately 20MB per sample in the batch.






\subsection{Synthesis from Hand-Drawn Streamlines}
\label{subsec:synthesis}

We now turn to our primary application: synthesizing entirely new vector fields from hand-drawn streamlines. In this task, the input streamlines (with constant magnitude) are sketched by the user.

As shown in Figure~\ref{fig:synthesis_results}, our diffusion model consistently produces more natural and physically plausible vector fields from hand-drawn inputs compared to the SVD baseline method. The figure demonstrates three distinct test cases: a simple vortex, a simple saddle, and a more complex S-shaped streamline that spans diagonally to create two vortex features with clear separation between them.

Our diffusion model offers flexibility through the Classifier-Free Guidance (CFG) parameter, which controls adherence to input streamlines. With low CFG values (top row), the model produces varied interpretations of simple inputs, while with CFG $\geq$ 1 (middle row), the model generates standard expected features that perfectly adhere to the input streamlines, resulting in a large simple vortex (left) and large simple saddle (middle) that closely match the conventional interpretation of these features.

Both configurations successfully generate flow features that blend naturally with the surrounding field, while the SVD method merely smooths the input streamlines without producing new features. This is evident in the S-shaped streamline case, where our model correctly interprets the user's intent to create two distinct vortex features, while SVD shows only a smoothed version with visible artifacts.

For best results, input guiding streamlines should be physically plausible, avoiding sharp turns, intersections, or merging of paths. When streamlines are hand-drawn, we recommend reducing the sampling rate of the input curve to smooth out minor inconsistencies from the drawing process, such as jitter. This will preserve the user's intended high-level structure while removing unintended fluctuations, as demonstrated with the S-shaped streamline (third column) in Figure~\ref{fig:synthesis_results}.

\section{Conclusion and Discussion}

We introduce a conditional diffusion model architecture specifically designed for vector field synthesis. We incorporate a classifier-free guidance to balance fidelity and plausibility of the synthesized fields. 
Our method offers some advantages over traditional optimization-based methods, including effective capturing of flow features like vortices and turbulent structures even with simple input, producing more physically authentic flow (\autoref{fig:synthesis_results}), and performing well across different types of flows (\autoref{fig:combined_results} and supplemental document).





Despite these advantages, our approach faces several limitations. First, diffusion models require more computational resources for training compared to traditional optimization methods. Second, the iterative denoising process during inference can be time-consuming, especially when using many diffusion steps for high-quality results.
Third, while our novel training approach enables vector field synthesis from sparse streamlines, 
generating coherent fields around complex input streamline constraints remains difficult. Specifically, when users provide streamlines with large variations or when datasets contain turbulent features with rapidly changing directions, the synthesized vector field may not smoothly adhere to input streamlines. 
Our current mitigation strategy involves discretizing input streamlines into smaller points, and we continue exploring enhanced training constraints to improve adherence.




Future work includes extending our approach to three dimensions, incorporating explicit physical constraints, exploring additional conditioning inputs, exploring more effective ways to specify the variations of the generated flows, and supporting interactive design for real-time vector field creation and editing. 

\acknowledgments{
This research was supported by NSF OAC 2102761.}

\bibliographystyle{abbrv-doi-hyperref}

\bibliography{Content/references}


\end{document}